\documentclass[runningheads]{llncs}
\usepackage[T1]{fontenc}
\usepackage{graphicx}
\usepackage[colorlinks=true, linkcolor=blue, urlcolor=blue, citecolor=blue]{hyperref}
\usepackage{xcolor}
\definecolor{customgreen}{HTML}{00A000} 
\usepackage{tcolorbox}
\usepackage{booktabs}
\usepackage{url} 
\usepackage{tabularx}
\usepackage{authblk}
\usepackage{pifont} 
\usepackage{afterpage}
\usepackage{enumitem}

\begin{document}
\pagestyle{plain}

\title{LLaVA-Pose: Enhancing Human Pose and Action Understanding via Keypoint-Integrated Instruction Tuning}
\titlerunning{Keypoint-Integrated Instruction-Following Data Generation}
%
\author{Dewen Zhang\inst{1}\orcidID{0009-0002-5145-0100} 
\and 
Tahir Hussain\inst{1}\orcidID{0009-0005-7937-6485} 
\and 
Wangpeng An\inst{2}\orcidID{0000-0002-1869-1837} 
\and 
Hayaru Shouno\inst{1}\orcidID{0000-0002-2412-0184}}
%
%
\institute{Department of Informatics, Graduate School of Informatics and Engineering, The University of Electro-Communications, Tokyo 182-8585, Japan\\
\email{\{zhangdewen,shouno\}@uec.ac.jp}\\
\email{(Tahir Hussain)f2240014@gl.cc.uec.ac.jp}\\
\and
TikTok Inc, 1199 Coleman Ave, San Jose, CA 95110\\
\email{anwangpeng@gmail.com}}
\maketitle              
\begin{abstract}
Current vision-language models (VLMs) are well-adapted for general visual understanding tasks. However, they perform inadequately when handling complex visual tasks related to human poses and actions due to the lack of specialized vision-language instruction-following data. We introduce a method for generating such data by integrating human keypoints with traditional visual features such as captions and bounding boxes, enabling more precise understanding of human-centric scenes. Our approach constructs a dataset comprising 200,328 samples tailored to fine-tune models for human-centric tasks, focusing on three areas: conversation, detailed description, and complex reasoning. We establish an Extended Human Pose and Action Understanding Benchmark (E-HPAUB) to assess model performance on human pose and action understanding. We fine-tune the LLaVA-1.5-7B model using this dataset and evaluate our resulting LLaVA-Pose model on the benchmark, achieving significant improvements. Experimental results show an overall improvement of 33.2\% compared to the original LLaVA-1.5-7B model. These findings highlight the effectiveness of keypoint-integrated data in enhancing multimodal models for human-centric visual understanding. Code is available at \url{https://github.com/Ody-trek/LLaVA-Pose}.

\end{abstract}
\section{Introduction}
The development of multimodal models integrating vision and language has become a central focus in artificial intelligence (AI) research~\cite{1,2,3,4}. Models like LLaVA (Large Language and Vision Assistant)~\cite{5} bridge the gap between visual perception and natural language understanding, utilizing large language models (LLMs) and visual encoders to process a wide range of image-related tasks~\cite{6,7,8}.

Despite their achievements, current models struggle with specialized tasks that require nuanced understanding of human activities, particularly those involving poses and actions. This limitation constrains their application in assistive robotics, healthcare, and human-computer interaction ~\cite{9,10,11,12}. A significant challenge is the lack of specialized vision-language instruction-following data. While LLaVA introduces a method for converting image-text pairs into instruction-following data using GPT-4~\cite{13}, it primarily relies on image captions and object bounding boxes, which lack the precision needed to interpret complex human activities. Consequently, models trained on such data show limited performance in tasks requiring detailed understanding of human pose and action.

Recent efforts have started to address this limitation by exploring language-guided pose understanding. For example, ChatPose~\cite{22} leverages LLMs to reason about 3D human poses from images and text, and PoseLLaVA~\cite{38} introduces a pose-centric multimodal LLM capable of unified pose estimation, adjustment, and generation. However, they primarily focus on SMPL-based~\cite{23} 3D representations, and do not explicitly target instruction-following tasks designed to facilitate interpretable human pose and action understanding in natural image scenarios.

To address this gap, we propose a novel approach that integrates human keypoints into the instruction-following data generation process. Our keypoint-integrated method provides a more comprehensive representation of human pose and action, enabling models to reason not just about objects in an image, but about how individuals interact with those objects and each other. This significantly enhances the model's ability to describe human movements, reason about their purposes, and respond to queries about human interactions.

Our contributions can be summarized as follows: 
\begin{sloppypar}
\begin{itemize}[label=\textbullet]
\item
We introduce a method for generating vision-language instruction-following data by integrating human keypoints, filling a critical gap in existing multimodal models for human-centric visual understanding.
\item
We demonstrate substantial improvements in human-centric visual tasks through comprehensive experiments comparing our fine-tuned LLaVA-1.5-7B model (hereafter referred to as LLaVA-Pose) with its original version and other models.
\item
We offer insights into how different types of fine-tuning data impact model capabilities for specific domains.
\end{itemize}
\end{sloppypar}

\section{Related Work}
\subsection{Instruction-Following Multimodal Models}
The LLaVA model~\cite{5} has made noteworthy progress by integrating vision encoders with LLMs to tackle various vision-language tasks. Similarly, models such as DeepSeek-VL2~\cite{33}, $V^*$~\cite{14}, Qwen2-VL~\cite{15}, InternVL3~\cite{34}, Janus-Pro~\cite{35}, VisionLLM~\cite{16} and Flamingo~\cite{17} have been developed for general vision understanding. Although effective for image description and elementary visual reasoning, these models aren't specifically designed for interpreting detailed human poses and actions. We introduce a method for generating instruction-following data specifically for human pose and action understanding by leveraging human keypoint information alongside traditional visual features. By integrating this enriched dataset into the fine-tuning process of the LLaVA-1.5-7B model~\cite{36}, we enhance its capacity for complex reasoning and detailed description of human-centric activities.

\subsection{Multimodal Human-Centric Visual Understanding}
Traditional human activity recognition typically relies on distinct models for specific tasks~\cite{18,19,20,21}, but recent research shows a trend toward unifying these capabilities within a single multimodal framework. For instance, ChatPose~\cite{22} uses LLMs to combine language-based reasoning with visual input for understanding and generating 3D human poses. PoseLLaVA~\cite{38} further advances this direction by introducing a pose-centric multimodal LLM that integrates a pose encoder-decoder into the LLaVA framework. It enables unified processing of pose estimation, pose adjustment, and pose generation tasks across pose, image, and text modalities. Moreover, PoseLLaVA introduces the PosePart dataset to complement existing pose-language datasets such as PoseFix~\cite{39} and PoseScript~\cite{40,41}, improving the model’s ability to perform fine-grained pose manipulation. Our approach similarly merges visual and language processing capabilities but differs by focusing on instruction-following data specific to human pose and action scenarios. Unlike ChatPose and PoseLLaVA, which utilize SMPL~\cite{23} parameters for 3D pose representation, our work remains in a 2D context while enhancing interpretative abilities through diverse instruction-following data. This design encourages the model to associate human pose and action cues with natural language queries and responses, facilitating more interpretable and explainable interaction in human-centric visual understanding tasks.

\section{Keypoint-Integrated Visual Instruction Data Generation}
\label{section 3}
\begin{sloppypar}
Large-scale multimodal datasets like LAION-5B~\cite{24}, CC-12M~\cite{25} and COYO-700M~\cite{26} have advanced vision-language models (VLMs). However, leveraging these datasets specifically for instruction-following tasks involving nuanced understanding of human pose and action remains underexplored.
\end{sloppypar}

Previous research such as LLaVA demonstrates promising results in generating visual instruction-following data using symbolic representations (captions and bounding boxes) to prompt language-only GPT-4~\cite{13}. Our approach enhances this foundation by integrating human keypoint data into the instruction-following data generation process. While LLaVA focuses primarily on captions and bounding boxes to represent visual content, our method enriches this representation by including annotations of keypoints, which capture precise positions and movements of human body parts within the scene.

\begin{figure*}[!t]
    \centering
    \includegraphics[width=\textwidth]{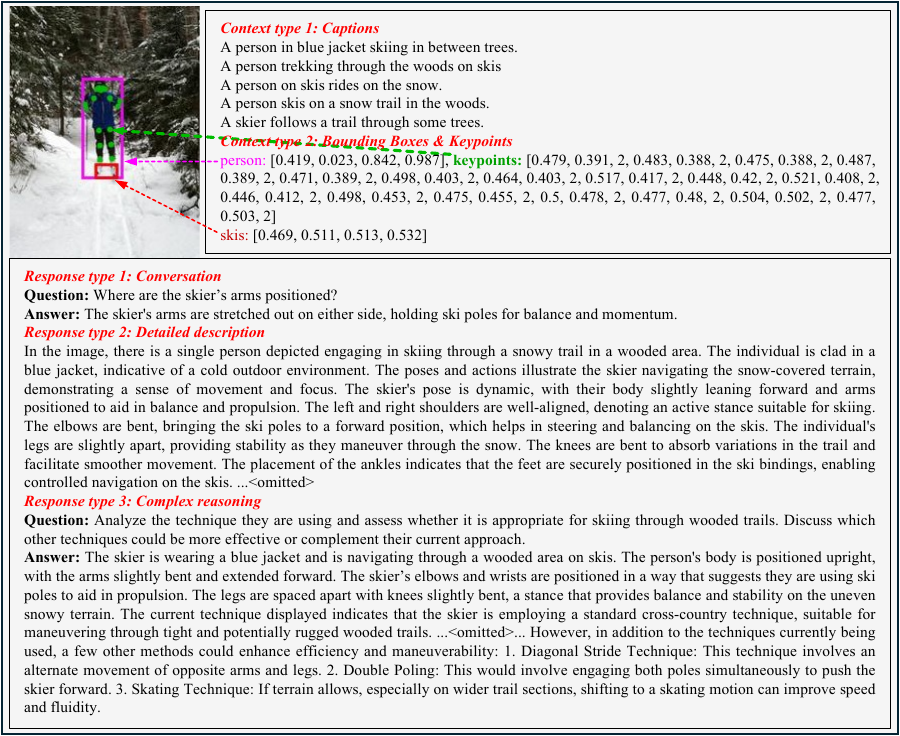} 
    \caption{One example to demonstrate the structure of instruction-following data. The top block displays the contexts information, including captions, bounding boxes (shown as solid rectangles in the visual image) and \textcolor{customgreen}{\textbf{keypoints}} (shown as \textcolor{customgreen}{\textbf{green}} circular markers in the visual image) used to prompt GPT-4o, and the bottom block displays the three types of responses generated. It is important to note that the visual image itself is not used to prompt GPT-4o, it is shown solely for reference purposes.}
    \label{figure 1}
\end{figure*}

To enhance the visual understanding capabilities of our model, we extend the data generation methodology originally used in LLaVA by incorporating human-centric annotations, using GPT-4o~\cite{27} as a teacher model. To represent images as visual features for prompting GPT-4o, our approach considers: (1) \textit{Captions} describing the visual scene from various perspectives; (2) \textit{Bounding boxes} localizing objects and providing spatial information; and (3) \textit{\textbf{Keypoints}} representing precise locations of joints and critical body parts. This enriched representation (example in top block of Fig.~\ref{figure 1}) allows for comprehensive understanding of human pose and action by detailing the exact positioning of body parts. The captions, bounding boxes, and keypoints annotations are obtained directly from the COCO dataset~\cite{28}.

Using these symbolic representations, we generate three distinct types of instruction-following data (example in bottom block of Fig.~\ref{figure 1}) from the COCO dataset~\cite{28}:

\begin{itemize}[label=\textbullet]
\item\textbf{Conversation:} Dynamic interactions simulating real-world conversational exchanges about human poses and actions, such as asking what individuals are doing in a given scene (see detailed prompts and curation process in Table~\ref{table 1}).
\item\textbf{Detailed description:} In-depth descriptions focusing on human body language and environmental interactions, transcending simple object identification to emphasize narrative-style outputs useful in applications requiring detailed human observation (see detailed prompts and curation process in Table~\ref{table 2}).
\item\textbf{Complex reasoning:} Challenges requiring multi-step reasoning about human activities, such as understanding intentions behind specific actions or predicting next possible movements based on current poses (see detailed prompts and curation process in Table~\ref{table 4}).
\end{itemize}

\begin{table*}[htbp]
    \caption{For each query, we demonstrate the process of building the prompt for GPT-4o to generate a multi-turn conversation about the image. The examples come from fewshot\_samples, where each example provides a short caption. We first construct few-shot demonstrations using context-response pairs, then query GPT-4o to generate new questions and answers focused on human poses and actions. \textbf{messages} form the final prompt.}
    \centering
    \begin{tcolorbox}[
        colback=gray!5, 
        colframe=black, 
        boxrule=0.3mm, 
        arc=0mm,
        sharp corners,
        boxsep=0.1mm
    ]
    \scriptsize
    \textbf{messages} = [\{\texttt{"}role\texttt{"}: \texttt{"}system\texttt{"}, \texttt{"}content\texttt{"}: f\texttt{"""}You are an AI visual assistant, and you are seeing a single image. What you see are provided with five sentences, describing the same image you are looking at. Answer all questions as you are seeing the image.

Design a conversation between you and a person asking about this photo. The answers should be in a tone that a visual AI assistant is seeing the image and answering the question. Ask diverse questions and give corresponding answers.

Your \textcolor{blue}{\textbf{primary focus}} should be on generating questions and answers about the \textcolor{blue}{\textbf{poses, actions, and movements of people}} in the image. Please generate diverse questions that relate primarily to \textcolor{blue}{\textbf{human poses and actions}}, and provide detailed answers as if you are seeing the image. Only include questions that have definite answers:\\
(1) one can see the content in the image that the question asks about and can answer confidently;\\
(2) one can determine confidently from the image that it is not in the image. Do not ask any question that cannot be answered confidently.
\texttt{"""}\}]
\\

\texttt{for sample in fewshot\_samples:} \\
\hspace*{2em}\textbf{messages}\texttt{.append(\{"role": "user", "content": sample["context"]\})} \\
\hspace*{2em}\textbf{messages}\texttt{.append(\{"role": "assistant", "content": sample["response"]\})} \\
\textbf{messages}\texttt{.append(\{"role": "user", "content": "Can you generate some new questions and answers focusing on the poses and actions of people in the image?"\})}
    
\end{tcolorbox}
\label{table 1}
\end{table*}

\begin{table*}[htbp]
    \caption{For each query, we demonstrate the prompt construction process for GPT-4o to generate detailed descriptions about human poses and actions. The examples come from annotations\_group, with each example containing an input annotation[``context'']. The final instruction in the prompt is randomly selected from \textbf{detailed\_description}, which consists of instructions listed in Table~\ref{table 3}. It is important to note that \textbf{messages} form the final prompt.}
    \centering
    \begin{tcolorbox}[
        colback=gray!5, 
        colframe=black, 
        boxrule=0.3mm, 
        arc=0mm,
        sharp corners,
        boxsep=0.1mm
    ]
    \scriptsize
    \textbf{messages} = [\{\texttt{"}role\texttt{"}: \texttt{"}system\texttt{"}, \texttt{"}content\texttt{"}: f\texttt{"""}You are an AI visual assistant specializing in analyzing human poses and actions in images. You receive five sentences, each describing the same image you are observing. In addition, specific object locations within the image are given, along with detailed coordinates. These coordinates are in the form of bounding boxes and \textcolor{blue}{\textbf{human keypoints}}, represented as (x1, y1, x2, y2) for bounding boxes and (x, y, visibility) for keypoints, with floating numbers ranging from 0 to 1. These values correspond to the top left x, top left y, bottom right x, and bottom right y for bounding boxes, and x, y coordinates along with visibility (0: not labeled, 1: labeled but not visible, 2: labeled and visible) for keypoints.

The keypoints represent the following body parts:\\
1. nose\\
2. left eye\\
3. right eye\\
4. left ear\\
5. right ear\\
6. left shoulder\\
7. right shoulder\\
8. left elbow\\
9. right elbow\\
10. left wrist\\
11. right wrist\\
12. left hip\\
13. right hip\\
14. left knee\\
15. right knee\\
16. left ankle\\
17. right ankle\\

Using the provided captions and bounding box/human keypoint information, \textcolor{blue}{\textbf{describe the scene in a detailed manner, focusing on the human poses and actions}}.

Instead of directly mentioning the bounding box or keypoint coordinates, utilize this data to explain the scene using natural language. Include details like the number of people, their \textcolor{blue}{\textbf{actions, poses, interactions, and relative positions}}.

When using the information from the caption and coordinates, directly explain the scene, and do not mention that the information source is the caption or the bounding box/keypoints. Always answer as if you are directly looking at the image.\texttt{"""}\}]
\\

\texttt{for annotation in annotations\_group:} \\
\hspace*{2em}\textbf{messages}\texttt{.append(\{"role": "user", "content": annotation["context"]\})} \\
\textbf{messages}\texttt{.append(\{"role": "user", "content": random.choice(}\textbf{detailed\_description}\texttt{)\})}

\end{tcolorbox}
    
\label{table 2}
\end{table*}

\begin{table*}[htbp]
\caption{The list of instructions for detailed image description about human poses and actions.}
\centering
\begin{tcolorbox}[
    colback=gray!5,
    colframe=black,
    boxrule=0.3mm,
    arc=0mm,
    boxsep=0.8mm,
    left=2mm, right=2mm, top=1mm, bottom=1mm,
    fontupper=\footnotesize,
    sharp corners
]
\ding{228} "Describe the actions and poses of people in the following image in detail." \\
\ding{228} "Provide a detailed description of the poses of people in the given image." \\
\ding{228} "Explain the various details of the actions of people you see in the image." \\
\ding{228} "Share a comprehensive analysis of the behaviors of people presented in the image." \\
\ding{228} "Offer a thorough analysis of the actions of people in the image." \\
\ding{228} "Explain the various poses and actions of people in the displayed image with great detail." \\
\ding{228} "Characterize the poses of people in the image using a well-detailed description." \\
\ding{228} "Break down the actions of people in the image in a detailed manner." \\
\ding{228} "Walk through the important details of the actions of people in the image." \\
\ding{228} "Portray the poses and actions of people in the image with a rich, descriptive narrative." \\
\ding{228} "Narrate the actions and poses of people in the image with precision." \\
\ding{228} "Analyze the poses and actions of people in the image in a comprehensive and detailed manner." \\
\ding{228} "Illustrate the actions and poses of people in the image through a descriptive explanation." \\
\ding{228} "Examine the actions and poses of people in the image closely and share their details." \\
\ding{228} "Write an exhaustive depiction of the actions of people in the given image." \\
\ding{228} "Carefully observe the people in the image and share the details of their poses and actions."
\end{tcolorbox}
\label{table 3}
\end{table*}

\begin{table*}[htbp]
    \caption{For each query, we demonstrate the prompt construction process for GPT-4o to generate complex reasoning responses about human poses and actions. The examples come from annotations\_group, with each example containing an input annotation[``context'']. It is important to note that \textbf{messages} form the final prompt.}
    \centering
    \begin{tcolorbox}[
        colback=gray!5, 
        colframe=black, 
        boxrule=0.3mm, 
        arc=0mm,
        sharp corners,
        boxsep=0.1mm
        ]
        \scriptsize
        \textbf{messages} = [\{\texttt{"}role\texttt{"}: \texttt{"}system\texttt{"}, \texttt{"}content\texttt{"}: f\texttt{"""}You are an AI visual assistant specializing in analyzing human poses and actions in images. You receive five sentences, each describing the same image you are observing. In addition, specific object locations within the image are given, along with detailed coordinates. These coordinates are in the form of bounding boxes and \textcolor{blue}{\textbf{human keypoints}}, represented as (x1, y1, x2, y2) for bounding boxes and (x, y, visibility) for human keypoints, with floating numbers ranging from 0 to 1. These values correspond to the top left x, top left y, bottom right x, and bottom right y for bounding boxes, and x, y coordinates along with visibility (0: not labeled, 1: labeled but not visible, 2: labeled and visible) for human keypoints.

        The human keypoints represent the following body parts:\\
        1. nose\\
        2. left eye\\
        3. right eye\\
        4. left ear\\
        5. right ear\\
        6. left shoulder\\
        7. right shoulder\\
        8. left elbow\\
        9. right elbow\\
        10. left wrist\\
        11. right wrist\\
        12. left hip\\
        13. right hip\\
        14. left knee\\
        15. right knee\\
        16. left ankle\\
        17. right ankle\\

        The task is to use the provided caption and bounding box/human keypoint information to create a plausible question about the human poses and actions in the image, and provide the answer in detail.

        Create \textcolor{blue}{\textbf{complex questions}} beyond describing the scene. To answer such questions, one should require first understanding the \textcolor{blue}{\textbf{human poses and actions}}, then based on the background knowledge or reasoning, \textcolor{blue}{\textbf{either explain why the actions are happening that way, or provide guidance and help to the user's request}}. Make the question challenging by not including the visual content details in the question so that the user needs to reason about that first.

        **Do not include any coordinates or numerical values in your explanation**. Instead, utilize the data to explain the scene using natural language. Include details like the \textcolor{blue}{\textbf{number of people, their actions, poses, interactions, relative positions, as well as the relationships and interactions between people and objects in the scene}}. Describe how people are using objects, their proximity to objects, and any activities involving both people and objects.

        When using the information from the caption and coordinates, directly explain the scene, and do not mention that the information source is the caption or the bounding box/human keypoints. Always answer as if you are directly looking at the image.
        \texttt{"""}\}]
        \\

        \texttt{for annotation in annotations\_group:} \\
        \hspace*{2em}\textbf{messages}\texttt{.append(\{"role": "user", "content": annotation["context"]\})}

    \end{tcolorbox}
    
    \label{table 4}
\end{table*}
\clearpage

\begin{sloppypar}
This approach generates 200,328 unique vision-language instruction-following samples (112,980 conversation, 45,174 detailed description, and 42,174 complex reasoning). These samples are specifically tailored to enrich the multimodal model's ability to interpret and engage with human pose and action understanding. For example, in scenarios involving skiing, as shown in Fig.~\ref{figure 1}, our approach uses keypoint data to provide nuanced understanding of the skier's posture, balance, and motion.
\end{sloppypar}

\section{Model Architecture and Fine-Tuning Approach}
Our LLaVA-Pose model adopts the simple yet powerful architecture of LLaVA-1.5~\cite{5}, which has demonstrated strong capabilities in visual instruction tuning while remaining highly data-efficient. We follow this architecture without structural modification, and focus on augmenting it with keypoint-integrated instruction-following data to enable improved understanding of human poses and actions.

Fig.~\ref{figure 2} provides an overview of our LLaVA-Pose model architecture:

\begin{figure}[h]
    \centering
    \includegraphics[width=\columnwidth]{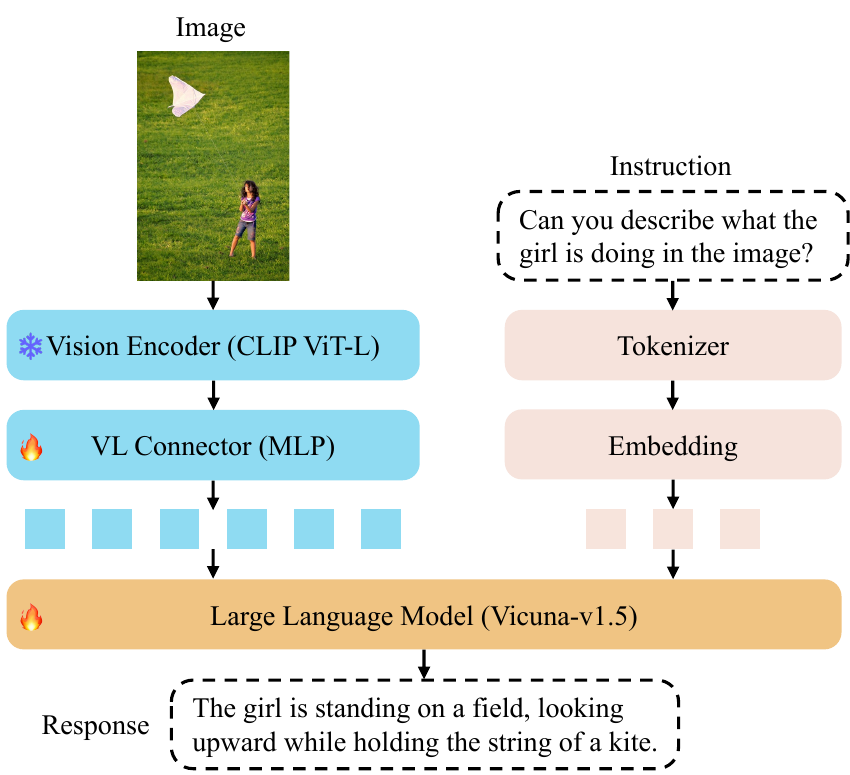}
    \caption{LLaVA-Pose model architecture.}
    \label{figure 2}
\end{figure}

\begin{itemize}[label=\textbullet]
\item\textbf{Vision Encoder:} The input image is first processed by a pre-trained CLIP ViT-L encoder~\cite{29}, which extracts visual features.
\item\textbf{Vision-Language Connector:} The visual features are passed through a two-layer multi-layer perceptron (MLP), which projects them into the embedding space of the LLM.
\item\textbf{Large Language Model:} The mapped visual embeddings are concatenated with the text token embeddings, and the combined embeddings are input into the LLM Vicuna-v1.5~\cite{30}. The LLM generates textual responses conditioned on both the visual and textual information.
\end{itemize}

\begin{sloppypar}
We fine-tune the entire model, comprising the MLP connector and the LLM, on our keypoint-integrated instruction-following data described in Sect.~\ref{section 3}. Through this fine-tuning process, the model enhances its ability to converse, describe and reason about regarding complex human-centric visual content. As shown in our experiments (Sect.~\ref{section 5}), this architecture significantly improves performance on human pose and action understanding tasks.
\end{sloppypar}

\section{Experiments}
\label{section 5}
We fine-tune the LLaVA-1.5-7B model to enhance its instruction-following ability in human pose and action understanding tasks, using a dataset of 200,328 unique samples generated from the COCO training dataset~\cite{28} by GPT-4o. The dataset consists of three instruction categories: conversation, detailed description, and complex reasoning (Sect.~\ref{section 3}). We refer to the resulting fine-tuned model as LLaVA-Pose. We train our model using the DeepSpeed framework~\cite{31} on 2×A100 GPUs, following hyperparameters similar to those used in the original LLaVA-1.5 model~\cite{5}, with adjustments for computational resources and stability, including a batch size of 8 and gradient accumulation steps of 2. Table~\ref{table 5} summarizes the hyperparameters used during fine-tuning.

\begin{table}[ht]
    \centering
    \caption{Fine-tuning hyperparameters used in our experiments.}
    \begin{tabular}{l|c} 
    \toprule
    Hyperparameter & Value \\
    \midrule
    batch size & 8 \\
    gradient accumulation steps & 2 \\
    learning rate & $2\times10^{-5}$ \\
    learning rate schedule & cosine decay \\
    learning warmup ratio & 0.03 \\
    optimizer & AdamW \\
    epoch & 1 \\
    weight decay & 0 \\
    DeepSpeed stage & 3 \\
    \bottomrule
    \end{tabular}
    \label{table 5}
\end{table}

\subsection{Qualitative Evaluation}

We conduct a qualitative evaluation to compare the responses of seven models: DeepSeek-VL2~\cite{33}, $V^*$~\cite{14}, Qwen2-VL-7B~\cite{15}, InternVL3-8B~\cite{34}, Janus-Pro-7B~\cite{35}, LLaVA-1.5-7B~\cite{36}, and our LLaVA-Pose. Table~\ref{table 6} and Table~\ref{table 7} present detailed comparisons of these models' outputs for two representative images from the COCO Validation dataset~\cite{28}, focusing on queries related to human pose and action understanding.

When asked to provide a detailed description of the poses and actions of the characters in the tennis player image (Table~\ref{table 6}), the responses from DeepSeek-VL2, $V^*$, Qwen2-VL-7B, InternVL3-8B, Janus-Pro-7B, and LLaVA-1.5-7B models offer simple descriptions, concentrating on general aspects of the scene. Their analyses remain relatively superficial, lacking detailed explanations of the characters’ postures, movements, positions, or interactions. In contrast, our LLaVA-Pose model delivers a more nuanced and contextually aware analysis. It identifies key elements of the main player's posture, including specific details such as the flexion of the knees, the positioning of the elbows and wrists, and the alignment of the shoulders. Additionally, our model clearly explains how these body parts contribute to the power and precision of the player's tennis swing. It also captures the passive involvement of two women in the scene, accurately describing their postures, attention levels, and engagement, which other models fail to recognize.

Similarly, when asked about the surfing image (Table~\ref{table 7}), LLaVA-Pose again outperforms the other models in providing a comprehensive and fine-grained analysis. While the baseline models typically state that the person is standing with bent knees and extended arms for balance, their explanations lack depth and contextual reasoning. In contrast, LLaVA-Pose offers a detailed description of the surfer’s posture and movement: it highlights the bent knees lowering the center of gravity, the extended arms functioning as a counterbalance, and how these pose elements contribute to agility, shock absorption, and control while maneuvering on the surfboard. Furthermore, it explains how the surfer's posture dynamically adapts to the changing conditions of the waves, which is an aspect not captured by the other models.

\begin{table*}[htbp]
    \caption{Comparison of responses from DeepSeek-VL2, $V^*$, Qwen2-VL-7B, InternVL3-8B, Janus-Pro-7B, LLaVA-1.5-7B, LLaVA-Pose models for the given tennis player image. \textcolor{blue}{Blue-highlighted} content indicates key information relevant to the question.}
    \centering
    \tiny
    \resizebox{\textwidth}{!}{  
    \begin{tabular}{p{\textwidth}}  
        \toprule
        {\textbf{A visual input example for the tennis player image}} \\
        \midrule
        \includegraphics[width=0.3\linewidth]{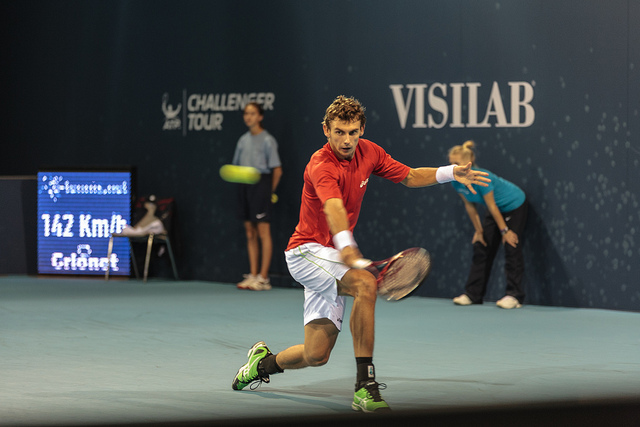} \\
        \tiny{Source:} \url{http://images.cocodataset.org/val2014/COCO_val2014_000000060992.jpg}
        \\
        \midrule
        User: Portray the poses and actions of people in the image with a rich, descriptive narrative. \\
        DeepSeek-VL2: The image captures a tennis player in action on a court during a match. The player is wearing a red shirt, white shorts, and green and white wristbands. He is in a tennis racket in his right hand, appears to be in the middle of a forehand shot. The player is in a fast-paced moment, with his body leaning forward and focused expression. In the background, there are two ball boys. One boy is holding a tennis ball, wearing a blue shirt and black shorts, while the other is wearing a blue shirt and black pants. They appear to be ball boys. The court has a blue wall in the background with the words "VISILAB" and "CHALLENEFR TOUR" written on it. There is also a digital scoreboard on the left side of the court displaying "142 Km/h" and "Cronet." The court is a professional tennis match, likely part of a Challenger Tour. \\
        \midrule
        User: Portray the poses and actions of people in the image with a rich, descriptive narrative. \\
        $V^*$: In the image, a man is swinging a tennis racket on a court, while two other people are present in the scene. One of the individuals appears to be a young girl, who might be watching the man playing tennis or waiting for her turn to participate. The man, wearing a red shirt, is actively engaged in the game, demonstrating his skills by swinging the racket with precision and power. It seems like an enjoyable and lively moment where the players share a common interest in tennis, possibly making it a family or social event where people of different ages can bond over the sport. \\
        \midrule
        User: Portray the poses and actions of people in the image with a rich, descriptive narrative. \\
        Qwen2-VL-7B: In the image, a tennis player is captured in the midst of an intense rally, showcasing his agility and skill. He is leaning forward, his body coiled like a spring, as he prepares to strike the incoming tennis ball with his racket. His eyes are focused intently on the ball, his muscles tensed and ready to spring into action. In the background, another person is holding a tennis ball, likely waiting for the right moment to serve or pass the ball to the player. This person's stance is relaxed yet attentive, as they watch the ongoing match with keen interest. A third person is standing nearby, possibly a coach or a fellow player, observing the game with a critical eye. Their presence adds an element of support and encouragement to the scene, as they watch the player's performance and offer guidance when needed. The image captures the dynamic and competitive nature of tennis, with each player fully engaged in the game and focused on their respective tasks. The vibrant colors of the court and the players' attire add to the overall energy and excitement of the scene. \\
        \midrule
        User: Portray the poses and actions of people in the image with a rich, descriptive narrative. \\
        InternVL3-8B: In the image, a man is playing tennis on a court, swinging his racket to hit a tennis ball. He is wearing a red shirt and white shorts, and his pose suggests that he is in the middle of a powerful swing. There are other people in the background, possibly watching the game or waiting for their turn to play. The scene takes place on a tennis court, which is a common setting for tennis matches and practice sessions. The presence of the other people and the tennis ball in the air indicate that this is an active and engaging moment in the game. \\
        \midrule
        User: Portray the poses and actions of people in the image with a rich, descriptive narrative. \\
        Janus-Pro-7B: In the image, a dynamic scene unfolds on a tennis court, capturing the intensity and focus of a player in action. The central figure, a tennis player dressed in a red shirt and white shorts, is mid-swing, his body leaning forward with a powerful motion. His right arm extends outward, gripping a tennis racket, while his left arm is slightly bent, maintaining balance. His legs are positioned in a wide stance, with the right leg slightly bent and the left leg extended backward, emphasizing the force and precision of his movement. The player's expression is one of concentration and determination, his eyes fixed on the ball as it approaches. The ball is suspended in the air, a blur of motion, indicating the speed and agility required in this moment. The player's grip on the racket is firm, his wrist steady, suggesting a practiced and skilled hand. In the background, two individuals stand observing the match. One, dressed in a light blue shirt and dark shorts, holds a tennis ball, ready to serve or return a shot. The other, wearing a teal shirt and black pants, leans forward slightly, perhaps in anticipation of the next move. Their presence adds a layer of context, indicating that this is a competitive or practice session. The backdrop of the court features a digital display showing the speed of the serve, with the text "142 km/h" and "Coulée" visible, adding a modern and technological touch to the scene. The wall behind the players is adorned with logos, including "VISIAB" and "CHALLENGER TOUR," suggesting a professional or high-level amateur event. The overall atmosphere is one of energy and focus, with each individual playing a role in the unfolding drama of the tennis match. The image captures a moment of athletic prowess and competitive spirit, highlighting the dedication and skill required in the sport. \\
        \midrule
        User: Portray the poses and actions of people in the image with a rich, descriptive narrative. \\
        LLaVA-1.5-7B: The image captures a tennis match in progress, with a man in a red shirt and white shorts playing the game. He is in the middle of a swing, holding a tennis racket and preparing to hit the ball. Another person is standing nearby, likely watching the game or waiting for their turn to play. There are a few more people in the scene, one of them holding a tennis ball, possibly waiting for their turn to play or observing the match. The presence of multiple people in the scene suggests that this is a social event or a gathering where people are enjoying a friendly game of tennis. \\
        \midrule
        User: Portray the poses and actions of people in the image with a rich, descriptive narrative. \\
        \textbf{LLaVA-Pose:} In the image, there are three individuals engaged in a dynamic scene on a tennis court. The primary focus is on a man who is actively \textcolor{blue}{swinging a tennis racket}, likely in the midst of a powerful serve or return shot. \textcolor{blue}{His body is positioned with a slight forward lean, indicating the force and precision of his movement}. His \textcolor{blue}{left shoulder is slightly raised}, and his \textcolor{blue}{right shoulder is lower}, suggesting the follow-through of a swing. His \textcolor{blue}{left elbow is bent}, and his \textcolor{blue}{right elbow is extended}, with his \textcolor{blue}{right wrist positioned to control the racket's direction}. His \textcolor{blue}{legs are spread apart}, with his \textcolor{blue}{left knee slightly bent} and his \textcolor{blue}{right knee more flexed}, providing a stable base for his powerful swing. To the left of the main player, there is a woman who appears to be observing the action. She stands upright with a \textcolor{blue}{relaxed posture}, her \textcolor{blue}{arms hanging naturally} by her sides. Her \textcolor{blue}{head is turned towards} the main player, indicating her focus on the ongoing play. Her \textcolor{blue}{stance is casual}, with her \textcolor{blue}{feet firmly planted} on the ground, suggesting she is not currently involved in the game but is instead watching attentively. Nearby, there is another woman who is also watching the game. She is \textcolor{blue}{positioned slightly behind} the main player, with her \textcolor{blue}{body turned towards} the action. Her \textcolor{blue}{posture is slightly more engaged} than the observer on the left, with her \textcolor{blue}{shoulders squared} and her \textcolor{blue}{head turned towards} the main player. Her \textcolor{blue}{arms are relaxed}, and her \textcolor{blue}{legs are slightly apart}, indicating a stance that is ready to move if needed. The scene captures the intensity and focus of a tennis match, with the main player in the midst of a powerful action, while the other two individuals are observing, possibly waiting for their turn to play or simply enjoying the game. The overall atmosphere is one of concentration and athleticism, with each person contributing to the dynamic environment of the tennis court. \\
        \bottomrule
    \end{tabular}
    }
    
    \label{table 6}
\end{table*}

\begin{table*}[htbp]
    \caption{Comparison of responses from DeepSeek-VL2, $V^*$, Qwen2-VL-7B, InternVL3-8B, Janus-Pro-7B, LLaVA-1.5-7B, LLaVA-Pose models for the given surfing image. \textcolor{blue}{Blue-highlighted} content indicates key information relevant to the question.}
    \centering
    \tiny
    \resizebox{\textwidth}{!}{  
    \begin{tabular}{p{\textwidth}}  
        \toprule
        {\textbf{A visual input example for the surfing image}} \\
        \midrule
        \includegraphics[width=0.3\linewidth]{COCO_val2014_000000004742.jpg} \\
        \tiny{Source:} \url{http://images.cocodataset.org/val2014/COCO_val2014_000000004742.jpg}
        \\
        \midrule
        User: Why is the person standing with knees bent and arms extended while riding the surfboard? \\
        DeepSeek-VL2: The person is standing with knees bent and arms extended while riding the surfboard to maintain balance. \\
        \midrule
        User: Why is the person standing with knees bent and arms extended while riding the surfboard? \\
        $V^*$: The person standing with knees bent and arms extended while riding the surfboard is likely doing so to maintain balance and control on the surfboard as it moves through the water. By bending the knees and extending the arms, the individual is able to increase the surface area of the body in contact with the water, which helps with stability. This posture is also known as the \"surf stance\" or \"paddling stance,\" and it allows the surfer to efficiently paddle and maneuver the board in the water before catching a wave. \\
        \midrule
        User: Why is the person standing with knees bent and arms extended while riding the surfboard? \\
        Qwen2-VL-7B: The person is standing with knees bent and arms extended while riding the surfboard to maintain balance and control while surfing. This stance allows the surfer to adjust their body position and weight distribution to navigate the wave effectively. By bending their knees and extending their arms, the surfer can shift their center of gravity and react to the changing dynamics of the wave, ensuring a smooth and controlled ride. This technique is essential for maintaining stability and preventing falls while surfing. \\
        \midrule
        User: Why is the person standing with knees bent and arms extended while riding the surfboard? \\
        InternVL3-8B: The person is standing with knees bent and arms extended to maintain balance and control while riding the surfboard on the wave. \\
        \midrule
        User: Why is the person standing with knees bent and arms extended while riding the surfboard? \\
        Janus-Pro-7B: The person is standing with knees bent and arms extended while riding the surfboard to maintain balance and stability. Bending the knees lowers the center of gravity, which helps in maintaining balance on the moving surfboard. Extending the arms can help with balance and provide additional stability. This posture is common among surfers to help them stay upright and maneuver the board effectively. \\
        \midrule
        User: Why is the person standing with knees bent and arms extended while riding the surfboard? \\
        LLaVA-1.5-7B: The person is standing with knees bent and arms extended while riding the surfboard to maintain balance and control on the surfboard. By bending their knees, they lower their center of gravity, which helps them stay stable on the surfboard and adapt to the changing water conditions. Extending their arms outward provides additional balance and stability, as well as helps them maintain a proper stance on the surfboard. This positioning allows the surfer to effectively maneuver the surfboard and ride the waves, ensuring a smooth and enjoyable surfing experience. \\
        \midrule
        User: Why is the person standing with knees bent and arms extended while riding the surfboard? \\
        \textbf{LLaVA-Pose:} The person is standing with knees bent and arms extended to maintain balance and stability while riding the surfboard. This stance is crucial for several reasons: 1. \textcolor{blue}{Balance and Stability:} Bent knees \textcolor{blue}{lower the center of gravity}, which helps the surfer stay balanced on the moving surfboard. This is particularly important when riding waves, as the board can be unstable and the water can be unpredictable. 2. \textcolor{blue}{Control:} Extended arms act as a \textcolor{blue}{counterbalance} to the body's movements. By spreading the arms, the surfer can make \textcolor{blue}{fine adjustments} to their balance, allowing them to \textcolor{blue}{shift their weight and direction} as needed to navigate the wave. 3. \textcolor{blue}{Shock Absorption:} The bent knees act as \textcolor{blue}{shock absorbers}, allowing the surfer to absorb the impact of the waves and maintain a smooth ride. This is especially important for beginners or those riding smaller waves. 4. \textcolor{blue}{Flexibility and Agility:} This stance allows the surfer to be more \textcolor{blue}{agile and responsive} to changes in the wave's shape and speed. It enables \textcolor{blue}{quick shifts in weight and direction}, which are essential for \textcolor{blue}{maneuvering the surfboard} effectively. In summary, the person's pose is a combination of balance, control, shock absorption, and agility, all of which are essential for successfully riding a surfboard on the waves. \\
        \bottomrule
    \end{tabular}
    }
    
    \label{table 7}
\end{table*}

\subsection{Quantitative Evaluation}
\begin{sloppypar}
To systematically evaluate performance, drawing inspiration from prior work~\cite{5,30}, we utilize GPT-4o~\cite{27} to measure response quality across different models. Following LLaVA's methodology, we create triplets (image, ground-truth descriptions, question) and have each model generate responses. A language-only GPT-4o then evaluates these responses on a scale of 1-10, considering helpfulness, relevance, accuracy, and detail level.
\end{sloppypar}

\subsubsection{Results.} To evaluate model performance on human-centric tasks, we construct the Extended Human Pose and Action Understanding Benchmark (E-HPAUB), an extended version of HPAUB proposed in our previous work~\cite{37}. In this extended version, we randomly select 90 images of human-centric scenes from the COCO Validation 2014 dataset~\cite{28} and generate three distinct types of questions for each image: conversation, detailed description, and complex reasoning, resulting in a total of 270 questions. The questions are crafted using the data generation approach outlined in Sect.\ref{section 3}. This extended benchmark enables a more comprehensive evaluation of the model's ability to interpret and respond to diverse human-centric visual scenarios involving human poses and actions. By systematically varying the training datasets, we analyze the impact of different types of instruction-following data on model performance, as shown in Table~\ref{table 8}. Results demonstrate substantial improvements compared to the original LLaVA-1.5-7B model:

\begin{itemize}[label=\textbullet]
\item
Conversation: The fine-tuned model scores 58.9 vs. original's 54.0
\item
Detailed description: 69.1 vs. 47.2
\item
Complex reasoning: 75.1 vs. 55.2
\item
Overall performance: The model fine-tuned on all three data types achieves 69.4 vs. the original LLaVA-1.5-7B's 52.1, representing a 33.2\% increase
\end{itemize}

\begin{sloppypar}
As shown in Table~\ref{table 8}, all models fine-tuned with keypoint-integrated instruction-following data outperform the baseline LLaVA-1.5-7B model that uses only captions and bounding boxes, demonstrating the significant contribution of keypoint information.
\end{sloppypar}

\begin{table}[htbp]
    \caption{Ablation study on E-HPAUB with various training data. We prompt GPT-4o to evaluate and compare responses generated by our LLaVA-Pose model against those from the original LLaVA-1.5-7B model. GPT-4o is tasked with assessing and providing ratings based on the overall quality of the answers, accompanied by detailed explanations.}
    \centering
    \normalsize 
    \resizebox{\textwidth}{!}{
    \begin{tabular}{l|c|c|c|c}
        \toprule
        & \textbf{Conversation} & \textbf{Detailed description} & \textbf{Complex reasoning} & \textbf{All} \\
        \midrule
        Full data & \textbf{64.3} & \textbf{78.9} & 65.0 & \textbf{69.4} \\
        Conversation & 58.9 & 45.1 & 50.4 & 51.5 \\
        Detailed description & 52.7 & 69.1 & 65.3 & 62.4 \\
        Complex reasoning & 62.4 & 59.7 & \textbf{75.1} & 65.7 \\
        LLaVA-1.5-7B~\cite{36} & 54.0 & 47.2 & 55.2 & 52.1 \\
        \bottomrule
    \end{tabular}
    }
    \label{table 8}
\end{table}

\begin{table}[htbp]
    \caption{Comparative performance of DeepSeek-VL2, $V^*$, Qwen2-VL-7B, InternVL3-8B, Janus-Pro-7B, and LLaVA-Pose on E-HPAUB.}
    \centering
    \small 
    \resizebox{\textwidth}{!}{
    \begin{tabular}{l|c|c|c|c}
        \toprule
        & \textbf{Conversation} & \textbf{Detailed description} & \textbf{Complex reasoning} & \textbf{All} \\
        \midrule
        DeepSeek-VL2~\cite{33} & 60.1 & 41.8 & 42.6 & 48.1 \\
        $V^*$~\cite{14} & 62.6 & 38.8 & 68.8 & 56.7 \\
        Qwen2-VL-7B~\cite{15} & 73.8 & 52.8 & 68.8 & 65.1 \\
        InternVL3-8B~\cite{34} & 75.4 & 56.2 & 67.1 & 66.3 \\
        Janus-Pro-7B~\cite{35} & 70.2 & 68.7 & 66.1 & 68.3 \\
        \textbf{LLaVA-Pose} & 77.4 & 58.6 & 72.9 & \textbf{69.6} \\
        \bottomrule
    \end{tabular}
    }
    \label{table 9}
\end{table}

Using the E-HPAUB benchmark, we further conduct a comparative quantitative evaluation of DeepSeek-VL2, $V^*$, Qwen2-VL-7B, InternVL3-8B, Janus-Pro-7B, and our LLaVA-Pose model. The results, summarized in Table~\ref{table 9}, demonstrate that LLaVA-Pose achieves the highest overall average score of 69.6, outperforming DeepSeek-VL2 (48.1), $V^*$ (56.7), Qwen2-VL-7B (65.1), InternVL3-8B (66.3) and Janus-Pro-7B (68.3). Notably, LLaVA-Pose surpasses the second-best model, Janus-Pro-7B, by an absolute margin of 1.3 points, corresponding to a relative improvement of 1.9\%. These results further demonstrate that even when compared against state-of-the-art VLMs in the multimodal understanding domain, our LLaVA-Pose model maintains a clear performance advantage in understanding and reasoning about complex human poses and actions.

\begin{sloppypar}
\subsubsection{Discussion.} Fine-tuning the LLaVA-1.5-7B model on keypoint-integrated instruction-following data significantly enhances its ability to understand human poses and actions. This fine-tuning process leads to substantial improvements in human-centric tasks, enabling multimodal AI systems to operate more effectively in real-world environments. Notably, our LLaVA-Pose model achieves the highest overall performance on the E-HPAUB benchmark. These results demonstrate the effectiveness of integrating keypoint-level information to improve multimodal understanding of complex human activities.
\end{sloppypar}

\section{Conclusion}
In this paper, we introduced a method for generating vision-language instruction-following data by integrating human keypoints alongside traditional captions and bounding boxes information. This keypoint-integrated approach significantly enhances a multimodal model’s ability to understand and reason about human poses and actions. Unlike the original LLaVA method, which primarily relied on coarse object-level annotations, our method leverages fine-grained spatial and structural cues from keypoints to improve interpretability in human-centric visual scenarios. Through rigorous experimentation and evaluation, our fine-tuned model, LLaVA-Pose, demonstrated superior performance across various tasks related to human pose and action understanding. These findings underscore the potential of integrating fine-grained human keypoint data to enhance the capabilities of multimodal AI systems. 

\subsubsection{Limitations and Future Work.} While LLaVA-Pose excels at understanding static images, it lacks temporal modeling capabilities, which limits its performance in tasks involving dynamic human actions and interactions. Recognizing the temporal evolution of poses is essential for applications such as video-based activity analysis and sports performance monitoring. Future work will explore extending our framework to sequential visual data by incorporating temporal modeling techniques such as recurrent networks or Transformer-based attention mechanisms. We believe this direction will further enhance the model’s ability to reason about human behaviors in complex, time-varying environments.

\subsubsection{Acknowledgements} This work was supported by JST SPRING, Grant Number JPMJSP2131.

%
%
%
\bibliographystyle{splncs04}
\bibliography{mybibliography}
\end{document}